\documentclass[11pt,letterpaper]{article}
\usepackage[letterpaper]{geometry}
\usepackage{acl2012}
\usepackage{times}
\usepackage{latexsym}
\usepackage{amsmath}
\usepackage{multirow}
\usepackage{url}

\usepackage{amsfonts}
\usepackage{graphicx}
\usepackage{xcolor}

\usepackage{booktabs}
\makeatletter
\g@addto@macro\bfseries{\boldmath}
\newcommand{\@BIBLABEL}{\@emptybiblabel}
\newcommand{\@emptybiblabel}[1]{}
\makeatother
\usepackage[hidelinks]{hyperref}

\usepackage[english]{babel}
\usepackage[autostyle, english=american]{csquotes}
\MakeOuterQuote{"}

\title{Polite Dialogue Generation Without Parallel Data}

\author{Tong Niu \and Mohit Bansal \\
  UNC Chapel Hill \\
  {\tt \{tongn, mbansal\}@cs.unc.edu}   }

\date{}

\begin{document}
\maketitle

\begin{abstract}
Stylistic dialogue response generation, 
with valuable applications in personality-based conversational agents, 
is a challenging task because the response needs to be fluent, 
contextually-relevant, as well as paralinguistically accurate.
Moreover, parallel datasets for regular-to-stylistic pairs are usually unavailable. 
We present three weakly-supervised models that can generate diverse polite (or rude) dialogue responses without parallel data.
Our late fusion model (Fusion) merges the decoder of an encoder-attention-decoder dialogue model with a language model trained on stand-alone polite utterances.
Our label-fine-tuning (LFT) model prepends to each source sequence a politeness-score scaled label (predicted by our state-of-the-art politeness classifier) during training, and at test time is able to generate polite, neutral, and rude responses by simply scaling the label embedding by the corresponding score.
Our reinforcement learning model (Polite-RL) encourages politeness generation by assigning rewards proportional to the politeness classifier score of the sampled response.
We also present two retrieval-based polite dialogue model baselines.
Human evaluation validates that while the Fusion and the retrieval-based models achieve politeness with poorer context-relevance, the LFT and Polite-RL models can produce significantly more polite responses without sacrificing dialogue quality.
\end{abstract}
\section{Introduction}
\label{sect:Introduction}
Generating stylistic, personality-based language is crucial to developing engaging, convincing, and trustworthy conversational agents, for their
effective application in intelligent tutoring, home assistance, online reservations/purchasing, health care, etc.
Most current chatbots and conversational models lack any such style, 
which can be a social issue because human users 
might learn biased styles from such interactions, e.g., kids learning to be rude because the dialogue system
encourages short, curt responses, and also does not itself use politeness to set an example.\footnote{https://qz.com/701521/parents-are-worried-the-amazon-echo-is-conditioning-their-kids-to-be-rude/}
In this work, we focus on the important and diverse paralinguistic style axis 
of politeness vs. rudeness~\cite{brown1987politeness}.

Generating stylistic dialogue responses is 
a substantially challenging task 
because the generated response needs to be syntactically 
and semantically fluent, contextually-relevant to the conversation, 
as well as convey accurate paralinguistic features.
This is further complicated by the fact 
that content and style are only 
available in separate unpaired datasets, 
as opposed to translation-type parallel datasets containing regular-to-stylistic text pairs. 
Hence, we need indirectly-supervised models 
that can incorporate style into the generated response in absence of parallel data 
(i.e., where the training data for the conversation versus style components comes from two different datasets or domains), while still maintaining conversation relevance.

In this work, we present three such weakly-supervised models\footnote{The first version of this paper with the three Fusion, Discrete-LFT, and Polite-RL models was submitted on Oct 1, 2017. The two retrieval baselines and the continuous version of the LFT model were added to the Feb 1, 2018 resubmission based on reviewer discussions.}
that can generate diverse, natural, and contextually-relevant polite (and rude) dialogue responses, 
using data from separate style and dialogue domains:
the \textit{Stanford Politeness Corpus}~\cite{Danescu-niculescu-mizil2013} 
with Wikipedia and StackExchange requests, 
and the \textit{MovieTriples Dialogue Corpus}~\cite{serban2016building} with IMSDB movie scripts, respectively.
Each of our three models is based on a state-of-the-art politeness 
classifier and a sequence-to-sequence dialogue model.
The first model (Fusion) employs a late fusion technique 
to merge the response generation decoder of the dialogue model 
with a language model trained on polite utterances 
chosen by the politeness classifier.
The second label-fine-tuning (LFT) model prepends to the input utterance a single politeness label whose embedding is continuously scaled by the politeness score of the target sequence during training. 
This score is determined by feeding the corresponding ground-truth target sequence to our politeness classifier.
During test time, we show that the LFT model is able to control the politeness level of generated responses by simply scaling the label's embedding by the continuous target politeness score of our choice.
Our third reinforcement-based model (Polite-RL) encourages politeness generation by using the continuous-scale politeness score 
of the decoder-sampled sentence as a reward (via mixed-objective policy gradient methods),
i.e., polite utterances are encouraged with positive reward, 
and rude ones discouraged with negative reward.

Hence, our models only need a style classifier (without parallel data) to automatically 
influence and encourage continuous-scale stylistic language generation in a complex dialogue setup, which also requires maintaining relevance 
to conversational context. Each of these models requires minimal changes 
to the architecture of either the underlying 
sequence-to-sequence (Seq2seq) dialogue base model or the style classifier, and hence can modularly 
update the architecture with the latest state-of-the-art dialogue models 
or style classifiers (and for diverse styles).
In addition, we also employ two retrieval-based models, where we output the response which has the highest match with the input context from a set of classifier-picked polite responses or manually-picked generic polite utterances.
These two retrieval models serve as parallel investigations on the performance of our three proposed generative models above.

We conducted multiple human evaluations (for style and dialogue quality) 
on Amazon Mechanical Turk 
(\textit{MTurk})~\cite{d548a2d3f3a54c6c9400ceeb66d003e7} for all three 
models plus the base sequence-to-sequence dialogue model and the retrieval-based models, and show that while the Fusion and the two retrieval models increase the politeness level of responses 
at the cost of poorer dialogue quality, both our LFT 
and Polite-RL models can successfully produce polite responses (capturing several politeness strategies discussed by~\newcite{brown1987politeness}), 
without sacrificing dialogue coherence and relevance  compared to the base Seq2seq model (hence better balance between politeness and dialogue quality). 
We also compare the output dialogue politeness levels of the continuous LFT model for three different politeness levels.
Finally, we present several detailed qualitative and quantitative analyses, including positive and negative output examples, automatic metric results on output responses, classifier error analysis, and visualization of the RL rewards.
\section{Related Works}
\label{sect:Related Works}

\subsection{Models for Style Transfer}
\label{subseect:Models for Style Transfer}

\paragraph{Style Transfer with Parallel Data}
There have been multiple works on style transfer with parallel data.
These tasks can often be solved by directly applying some variation 
of translation-based Seq2seq model discussed in the previous section.
For example,~\newcite{xu2012paraphrasing} use a phrase-based statistical model, and~\newcite{jhamtani2017shakespearizing} use a standard Seq2seq model
to convert modern language to Shakespeare-style language 
by treating style transfer as a translation task.
Some labeled sequence transduction methods have also been proposed~\cite{kobus2016domain,yamagishi2016controlling,johnson2016google}. For example,~\newcite{DBLP:journals/corr/KikuchiNSTO16}
are able to control the length of the summarization text 
by feeding to the Seq2seq base model a label 
that indicates the intended output length 
in addition to the source input.
Our LFT model also adopts this labeling idea, and is able to handle a similar situation but without parallel data, because by labeling each target sequence in the training set with its politeness classifier score, we are essentially converting non-parallel data to (noisy) parallel data (by using a classifier with high accuracy).

\paragraph{Style Transfer without Parallel Data}
Several previous works have looked at style transfer without parallel data, in both vision~\cite{gatys2016image,zhu2017unpaired,liu2016coupled,liu2017unsupervised,taigman2016unsupervised,kim2017learning,yi2017dualgan},
and text~\cite{Sennrich2016a,hu2017controllable,DBLP:journals/corr/GhoshCLMS17,zhao2017learning,mueller2017sequence,wang2017steering,luan2017multi}.
Among these models, some are bag-of-words based,
i.e., they use style-related keywords to annotate 
the target sequences in the training set.
For example, to control how formal the output sequences are
in a EN-DE translation task,
~\newcite{Sennrich2016a} labeled each target sequence
based on whether it contains formal or informal verbs and pronouns (honorifics).
To build a language model that generates utterances with the desired style,~\newcite{ficler2017controlling} annotated their text with meta-data and keywords/POS tags based
heuristics, while~\newcite{DBLP:journals/corr/GhoshCLMS17}
also adopted keyword spotting based on a dictionary of emotional words.
The basic ideas of their models are similar to that of our LFT model. 
However, these keyword-spotting approaches do not fully extend to our politeness generation task,
because politeness strategies follow complex patterns of grammar, word order, and phrasing~\cite{Danescu-niculescu-mizil2013}.
For example, the politeness of \textit{please} 
depends on where it occurs in a sentence, 
and what other politeness markers it co-occurs 
with (e.g., `could/would you' style counterfactual modals vs. `can/will you' style indicative modals).
Therefore, our novel polite dialogue models are
based on an accurate neural classifier, 
which is better at capturing several compositional paralinguistic features (as visualized in~\newcite{Aubakirova2016}, 
whose politeness classifier we extend). 
Moreover, our LFT and Polite-RL models can generate a continuum of style levels based on the continuously-scaled (by the politeness score) label embedding or reinforcement rewards.

Lastly, there have also been style transfer models that rely on the latent representation 
of text and use variational auto-encoders or cross-alignment to disentangle the representation of content and style in text~\cite{hu2017controllable,shen2017style,zhao2017learning,fu2017style}.
During inference time, the latent style representation is combined
with new content to generate stylized, content-preserving text.
Although both fall into the category of style transfer, our task differs in two important aspects from their tasks.
First, as opposed to the task of strict content preservation when rephrasing a sentence to a different style, our task is about maintaining good relevance to the context when adding style, especially useful for dialogue-based tasks.
Another distinctive trait of our task is that politeness resides in a spectrum rather than a fixed category or topic (e.g., Shakespearean), 
and our models can treat politeness as a continuum, i.e., controlling the politeness level by adjusting the fusion rate in the Fusion model, the magnitude of the continuous label in the LFT model, or the RL weight in the Polite-RL model.

\subsection{Multi-Task Learning and Style Transfer}
\label{subsect:MTL}
In order to obtain a persona-based conversational agent,~\newcite{luan2017multi} proposed a multi-task learning (MTL) based approach: they train a Seq2seq model with conversation data and an autoencoder with non-conversational persona-related data from target speakers, and share the decoder parameters of these two models so that the generated responses can be adapted to the style of the target-speaker. 
This way of incorporating MTL into Seq2seq learning was first investigated by~\newcite{dong2015multi} and~\newcite{luong2015multi} to achieve multilingual NMT. In addition,~\newcite{sennrich2015improving} also employed MTL to improve NMT models with monolingual (non-parallel) data.
These approaches are related to our Fusion model,
because we use our classifier to obtain noisy polite target sequences (non-parallel data) which a polite language model trains on, and during inference combine the parameters of the language model with a generative dialogue model trained on parallel data.
In general, our models are also related to previous works like~\newcite{johnson2016google}, who adopted labeled sequence transduction methods for MTL tasks, because our task also involves adapting generated responses to different politeness styles and optimizing two sub-tasks' 
(namely response and politeness generation) loss functions (related to a multi-task setup).

\subsection{Politeness Studies}
\newcite{Danescu-niculescu-mizil2013} created the Stanford Politeness Corpus and trained an SVM classifier using a list of useful linguistic features based on strategies from Brown and Levinson's \textit{theory of politeness}~\cite{brown1987politeness}.
~\newcite{Aubakirova2016} recently took an end-to-end neural approach to this politeness classification task by training a CNN model that directly learns to identify polite requests
without using any hand-engineered features,
while still improving on prediction accuracy. They also visualized what features the CNN model was learning and discovered some new features along the way.
Our classifier mainly extends their work by
adding a bi-directional LSTM layer~\cite{hochreiter1997long,schuster1997bidirectional} before the CNN layer to capture long-distance relationships in the sentence, which leads to higher cross-domain performance.

A related early work in personality-based dialogue is~\newcite{mairesse2007personage}, who study introvert/extrovert personality language based on templated content and sentence planning (via personality dimensions such as hedges, tag questions, negations, subject implicitness, etc.). Relatedly,~\newcite{Sennrich2016a} use an English to German translation task to present a model that can generate target sequences that are either formal or informal, specifically based on honorifics-related verbs and pronouns. Our task is more general, taking into account several politeness-related paralinguistic features of~\newcite{brown1987politeness} and allowing end-to-end trainable stylistic dialogue generation with a polite-to-rude spectrum (based on a politeness classifier, without relying on parallel data). Moreover, our approaches allow simply replacing the politeness classifier with any other emotion or personality based language classifier to generate stylistic dialogue for that new style dimension.

\section{Politeness Classification Model}
\label{sect:Politeness Classification Model}

\begin{figure}[t]
\centering
\includegraphics[width=0.35\textwidth]{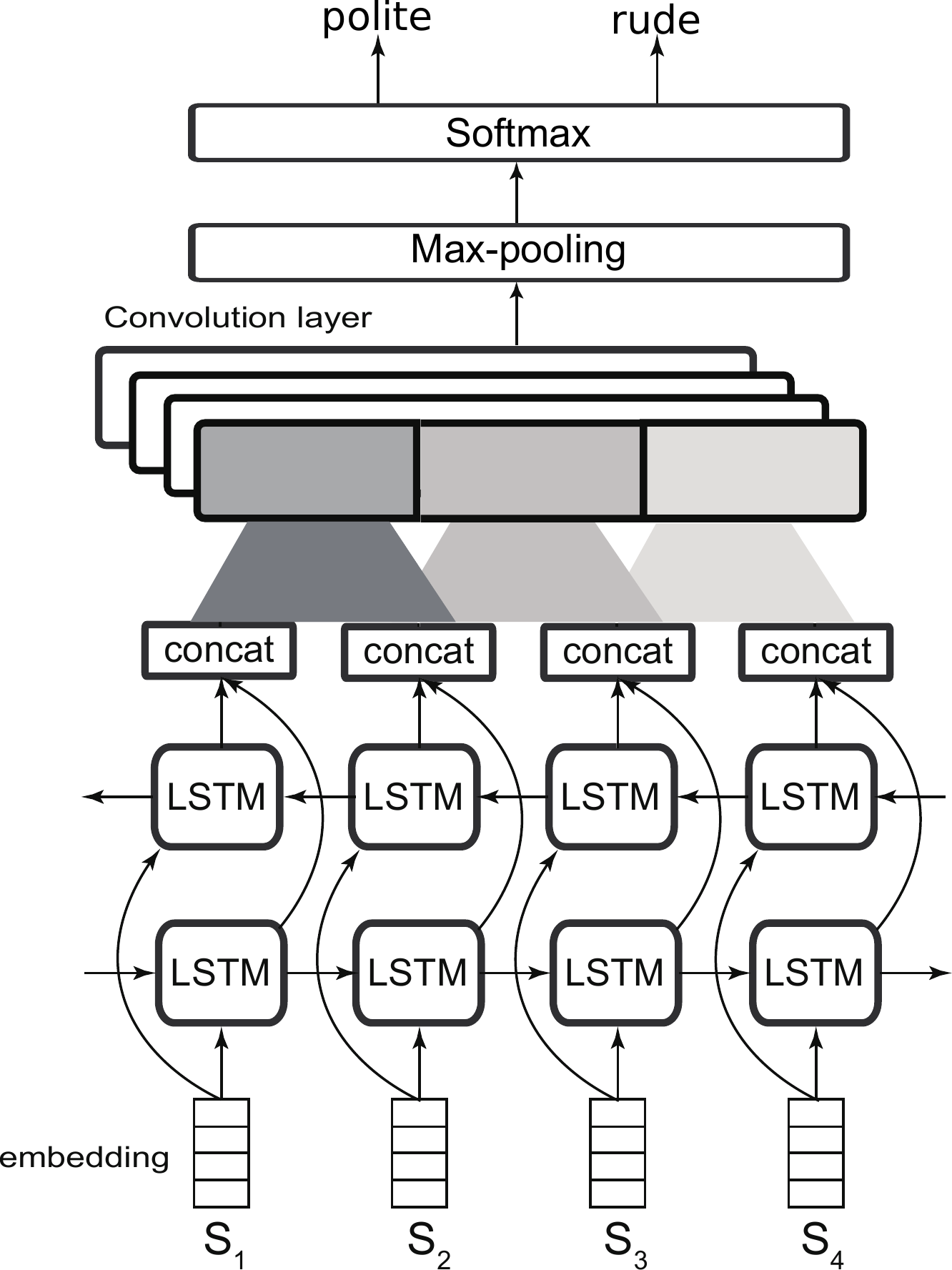}
\vspace{-7pt}
\caption{Our LSTM-CNN politeness classifier.
}
\label{fig:classifier}
\vspace{-7pt}
\end{figure}

In order to develop an accurate politeness classifier 
for effective use in stylistic dialogue response generation, 
we extend and improve upon the state-of-the-art CNN model 
of~\newcite{Aubakirova2016}, and propose
a bi-directional LSTM followed by a convolutional layer (see Figure~\ref{fig:classifier}), in order to both capture long-distance relationships in the sentence as well as windowed filter based features.
For a sentence $v_{1:n}$ 
(where each token $v_i$ is a $d$-dim word embedding vector), 
the LSTM layer first produces hidden states $h_{1:n}$ 
(where $h_t$ is the concatenation of forward 
and backward hidden states at time step $t$). 
A filter $m$ is then applied on a window of $u$ hidden states. 
This produces a convolution feature $c_i = f(m * v_{i:i+u-1} + b)$, 
where $f$ is a non-linear function and $b$ is a bias term. 
Every feature map $c \in \mathbb{R}^{n-u+1}$ is applied to each window,
so that $c = [c_1, ...,  c_{n-u+1}]$. 
The output of the convolutional layer is then 
fed to a max-pooling layer~\cite{DBLP:journals/corr/abs-1103-0398}
which gives $C = \max \{ c \}$ for the filter.
Filters of various sizes are used to obtain multiple features.
The result is then passed to a fully-connected softmax layer that
outputs probabilities over two labels, namely \textit{Polite} and \textit{Rude}.

Our classification model achieves comparable in-domain accuracy 
and improved cross-domain accuracy 
over the state-of-the-art results reported in~\newcite{Danescu-niculescu-mizil2013}
and~\newcite{Aubakirova2016}.
We will discuss these results in detail
in Section~\ref{sect:Results}.

\begin{figure}[t]
\centering
\includegraphics[width=0.48\textwidth]{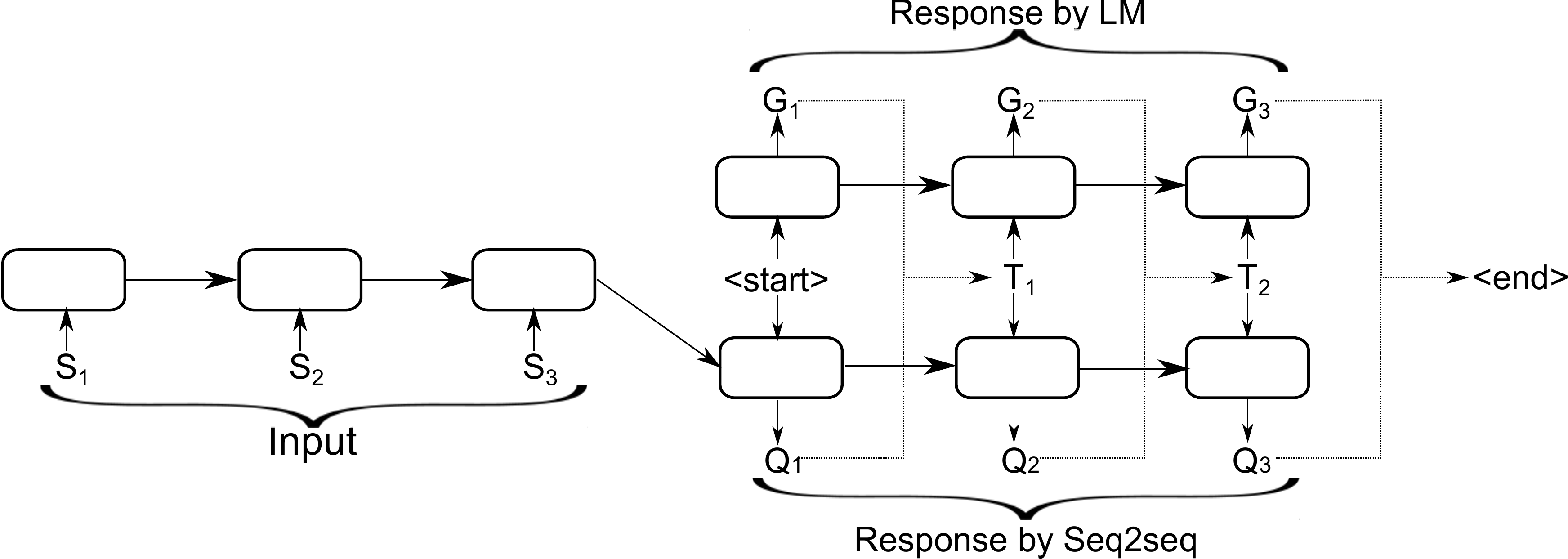}
\caption{Fusion model: the output probability distributions of the decoder and the polite-LM are linearly mixed to generate the final decoded outputs.}
\label{fig:fusion-model}
\end{figure}

\section{Polite-Style Dialogue Models}
\label{sect:polite-dialogue-models}
In this section, we first describe our base dialogue model, i.e., the core (backbone) dialogue architecture upon which the three proposed politeness models are built, and then present these three models that can generate polite dialogue responses. As a parallel investigation on the performance of our proposed models, we also employ two retrieval-based polite dialogue models toward the end.

\subsection{Base Seq2seq Dialogue Model}
\label{subsect:Base Seq2seq Dialogue Model}
Our base dialogue model is a simple 
sequence-to-sequence (Seq2seq) model that consists of a two-layer 
bi-directional LSTM-RNN encoder to encode the conversation history turns,
and a four-layer LSTM-RNN decoder to generate the response.
Additive attention from the output of the encoder
is applied to the last layer of the decoder.
This architecture is almost identical to that proposed
by~\newcite{Bahdanau2015}, except with more layers (similar to~\newcite{shao2017generating}).
Our base dialogue model achieves perplexity 
and word error rate results on par with those reported 
for the popular hierarchical HRED 
models in~\newcite{serban2016building}, thus serving as a good base model to incorporate style into. Details will be discussed in Section~\ref{sect:Results}.

\subsection{Fusion Model}
Inspired by the `late fusion' approach 
in~\newcite{DBLP:journals/corr/VenugopalanHMS16}, 
our Fusion model (Fig.~\ref{fig:fusion-model}) combines the 
response generation decoder of the base Seq2seq dialogue model with a language model (polite-LM) trained 
exclusively on polite utterances.
These utterances are chosen by feeding to the classifier 
all response utterances in the MovieTriples training set, 
and only keeping those with politeness scores 
great than a certain threshold 
(set to $0.8$ in our experiments, as will be discussed 
in Section~\ref{subsect:Retrieval-based Models}).
The polite-LM model is a two-layer LSTM-RNN based on~\newcite{DBLP:journals/corr/JozefowiczVSSW16}.

During inference time, 
we used the language model to re-score the final output of the Seq2seq decoder (for each time step)
by computing a linear combination of the output vocabulary distributions proposed by the Seq2seq model and polite-LM.
Specifically, let $p_t^{\textsc{S2S}}$ and $p_t^{\textsc{LM}}$ denote 
the output probability distributions proposed 
by the Seq2seq model and the LM model at time $t$, respectively.
The final `fused' distribution $p_t$ for that time step is:
\begin{align}
\label{eq:fusion}
	p_t &= \alpha \, p_t^{\textsc{S2S}} + (1 - \alpha) \, p_t^{\textsc{LM}}
\end{align}
where the \textit{fusion ratio} $\alpha$ is a hyperparameter
that indicates how much Seq2seq output
will influence the final output.

\begin{figure}[t]
\centering
\includegraphics[width=0.48\textwidth]{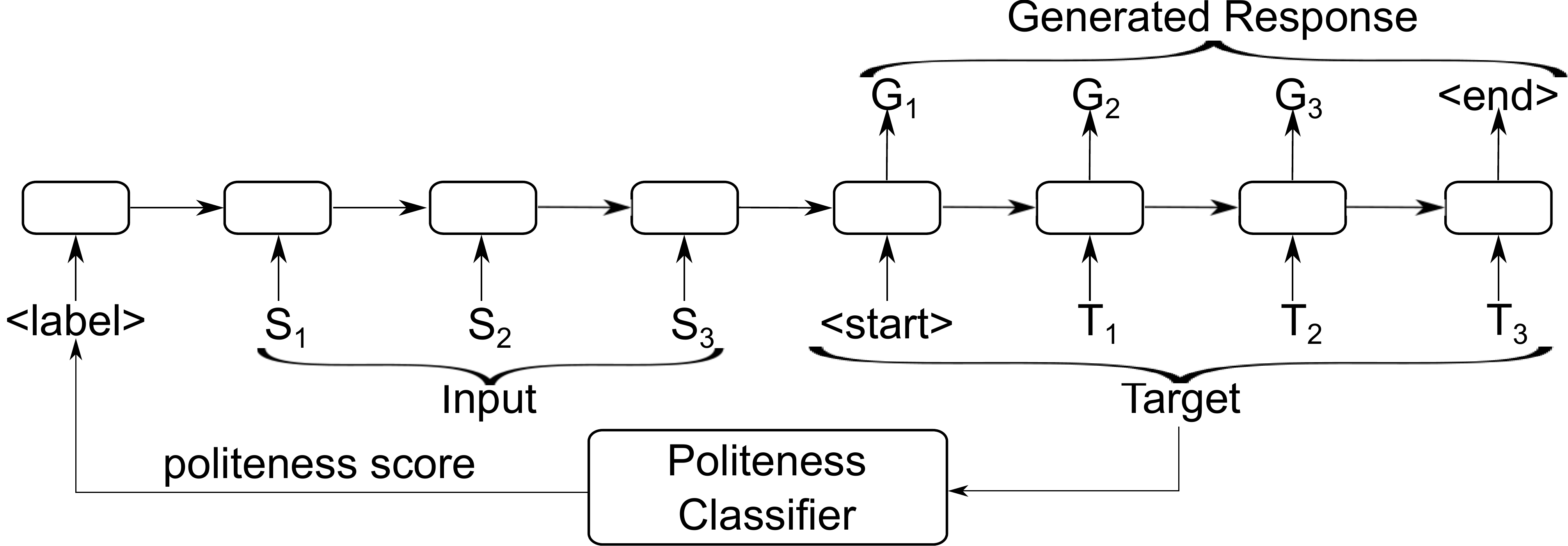}
\caption{\label{fig:label-model} Label-Fine-Tuning model: during training, the embedding of the prepended label is scaled by the style classifier's continuous score on the ground-truth (target) sequence. During testing, we scale the embedding of the label by the desired (continuous) politeness score of the generated response.
}

\end{figure}
\begin{figure*}[t]
\centering
\includegraphics[width=0.9\textwidth]{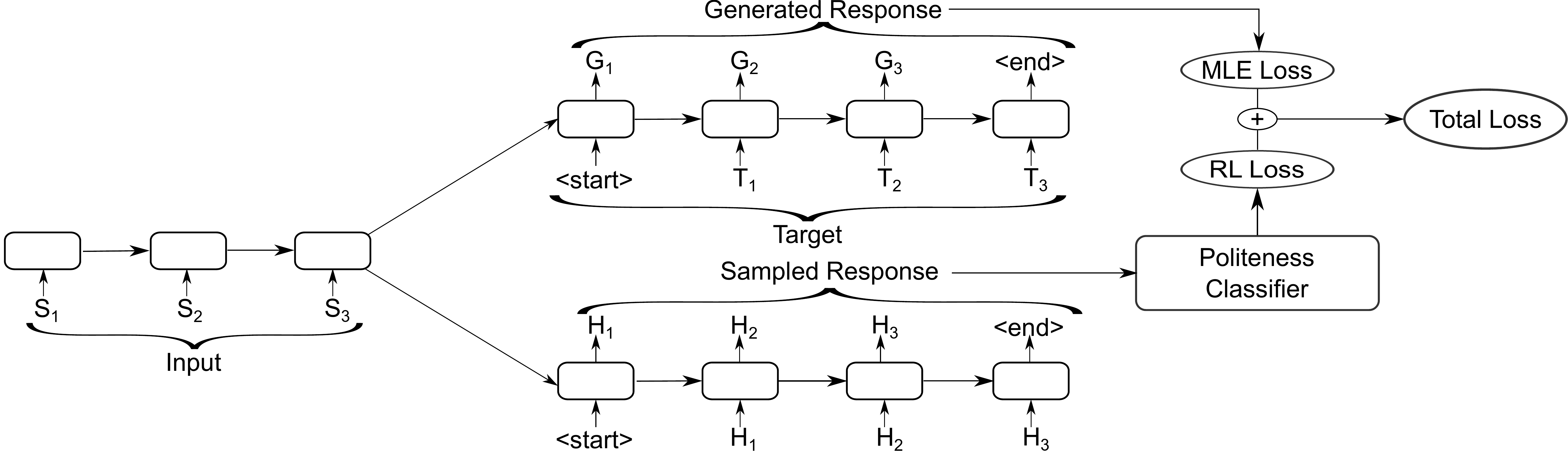}
\caption{Polite-RL model: upper-right shows max-likelihood (ML) training with generated and ground-truth target sequences; lower-right shows RL training with a randomly sampled response generated by the model and the reward it generates after getting fed into the style classifier. Note that the attention mechanism is not shown here for clarity.}
\label{fig:RL-model}
\end{figure*}

\subsection{Label-Fine-Tuning Model}
\label{subsect:Label-Fine-Tuning (LFT)}
There are at least two drawbacks of the Fusion model. 
First, half of its output is determined by a polite language model
that has not attended to the conversation context, making the response more likely to be irrelevant.
Second, the model does not learn politeness during training, but is forced to be polite only during inference time.
To address these two issues, we present our label-fine-tuning (LFT) model, which prepends a predicted continuous style label at the beginning of each input sentence 
to specify the intended politeness level. 

Specifically, we add to the vocabulary a single politeness label and attach with it a trainable word embedding, 
just like what we would do to a normal token. 
Then, the way we make it continuous is by scaling its embedding vector with the (intended) politeness score of the target sequence.
During training, this score is obtained 
by feeding the ground-truth target sequence (response) 
to the politeness classifier (see Figure~\ref{fig:label-model}), 
while during test time, 
we are free to scale the prepended politeness label with different scores of our choice (i.e., when we want the model to generate a polite response, we scale the label's embedding by a score between $0.5$ and $1.0$, while to generate a rude response, we scale the embedding by a score between $0.0$ and $0.5$).
This approach is related to the `numerically-grounded' language model~\cite{spithourakis2016numerically}, 
except that we scale the politeness label embedding by its corresponding politeness score, 
rather than concatenating the two as input to the LSTM.\footnote{Although we trained the politeness classifier to be binary, 
its outputs are probabilities ranging from $0.0$ to $1.0$.
This allows us to interpret the outputs 
as continuous politeness scores.}

Thus, the LFT model is able to simultaneously produce polite, neutral and rude responses depending on the prepended label, similar to recent multi-label, multi-space, and zero-shot machine translation work using language identity or style labels~\cite{Sennrich2016a,johnson2016google,DBLP:journals/corr/GhoshCLMS17}.
Intuitively, this prepended label serves as the prior 
for the intended style of the generated response sequence,
while the source utterance serves as the prior
for the content of the generated sequence.
In other words, the label and the source sentence
cooperatively determine what the overall response looks like.\footnote{Note that the position of the label did not affect the results much (e.g.,~\newcite{Sennrich2016a} appended the label
at the end of the input sequence). Moreover, our models use a bidirectional encoder, which does not distinguish between the beginning and end of the source sequence.}

\subsection{Polite Reinforcement Learning Model}
The LFT model incorporates style more directly into its training procedure than the fusion model, but it still does not fully exploit the value of the style classifier since it only supervises the dialogue model once by initially classifying the style of all the target sequences in the training set. Ideally we would want the classifier to constantly monitor and influence what style the model produces. Moreover, many contexts do not naturally elicit a polite response,\footnote{For example, it is hard to be polite in answering questions like "What's your name?" (The most "legitimate" answer would be "My name is XX.", rather than "Thanks for asking! My humble name is XX if you would allow me to say so.")} in which case we do not want to force the model to generate an utterance that matches the target politeness score, but rather to ask the model to generate as polite and natural a response as it could.
These limitations motivate us to propose the third model: Polite Reinforcement Learning model (Polite-RL), where the style classifier regularly updates the model parameters (via sampling-based policy gradient) with continuous-spectrum rewards that encourage decoder-generated response samples to be polite and discourage them from being rude.

Following work from~\newcite{DBLP:journals/corr/PaulusXS17},
our loss function consists of two terms.
The first term is the traditional maximum likelihood loss ($L_{\textsc{ml}}$),
which we refer to as the \textit{teacher forcing part}.
The other one is the reinforcement learning loss ($L_{\textsc{rl}}$)
based on politeness scores,
which we refer to as the \textit{reinforce part}.
The total loss $L$ then takes the form:
\begin{align}
\label{eq:LFT}
	L &= L_{\textsc{ml}} + \beta \, L_{\textsc{rl}}
\end{align}
where $\beta$ is a hyperparameter
indicating how much weight we want to give to the style reward component of the loss.
The teacher forcing part minimizes the average 
of the maximum-likelihood loss at each decoding step. 
Specifically, let $y^* = \{y_1^*, y_2^*, ..., y_n^*\}$ be the ground-truth response 
for a given source (conversation history) utterance sequence $x$. 
The maximum-likelihood training objective is the minimization of the loss:
\begin{align}
	L_{\textsc{ml}} &= - \sum^n_{t=1} \log p(y_t^* | y_1^*, ..., y_{t-1}^*, x)
\end{align}

We use a policy gradient method~\cite{williams1992simple,Sutton00policygradient} to calculate the second term 
in the objective function.
Specifically, we sample a generated response for each input sequence (conversation history) $x$,
and assign to it a reward $R$, 
which in our case is the politeness classifier's probability of the response classified as polite.
Let $y^s = \{y_1^s, y_2^s, ..., y_n^s\}$ be the sampled response,
then the reinforce part of the loss is:
\begin{align}
\label{eq:reinforce}
	L_{\textsc{rl}} = - \, (R - R_b) \, \sum^n_{t=1} \log p(y_t^s | y_1^s, ..., y_{t-1}^s, x)
\end{align}
where $R_b$ is a baseline that helps reduce variance during training~\cite{ranzato2015sequence}.

Note that we can invert the classifier scores or reward (by flipping the first minus sign in Eq.~\ref{eq:reinforce}), if we want to encourage rudeness as the style, instead of politeness.
This also shows that an advantage of our implementations of the LFT model over the Polite-RL model (at the cost of shallower training) is that the LFT model can multitask to simultaneously produce responses of different style labels at test time, whereas reward-based reinforcement learning can only work in one direction at a time (based on the reward sign).\footnote{However, to make the reward-based model capable of multitasking, one could also prepend various politeness labels to each of the context in the training set (thus generating several examples out of one context), and encourage the generated response to be consistent with the given label. We will explore this extension in future work.}

\subsection{Retrieval-based Models}
\label{subsect:Retrieval-based Models}
We employ two retrieval-based baseline models 
as a sanity check to the proposed approaches' performance: the first with oracle-level fluency, the second with additional oracle-level politeness.

\paragraph{Classifier-based Retrieval}
Following~\newcite{Lowe2015}, 
for a $[X_1, Y, X_2]$ triple, our retrieval model treats the context ($X_1, Y$)
and each response ($X_2$) as two documents and convert them to their TF-IDF based vectors~\cite{ramos2003using} to check for similarity.
Specifically, we first obtain all candidate responses 
in the training set that are polite,\footnote{We treat only responses in the higher, more-accurate percentile of $[0.8, 1.0]$ range as \textit{polite} (and $[0.0, 0.2]$ range as \textit{rude}).} and calculate their TF-IDF vectors.
Then for each context TF-IDF vector in the test set,
we calculate its cosine similarity with that of each such polite-classified candidate response, 
and output the one with the highest value.
Intuitively, for each context we are choosing a response that is both polite and most relevant to (having the most word overlaps with) the context.

\paragraph{Generic-$10$}
This approach is similar to the one above but uses the $10$ manually-chosen most-polite generic  utterances as candidate responses for each context.
Specifically, we collect all ground-truth polite requests from 
the Stanford Politeness Corpus, split each one into sentences, and then manually pick the most frequent $10$ polite sentences.\footnote{The $10$ final polite sentences for Generic-$10$ are "thanks.", "can you help?", "can you clarify?", "no problem.", "you're welcome.", "interesting question.", "thanks for the answer.", "could you help please?", "can you elaborate?" and "nice.". The $2$ rejected ones are "what have you tried?" and "what do you think?". This shortlist needed some human filtering because in the Stanford Politeness Corpus, each polite example consists of two sentences, and sometimes not both of them are polite, i.e., one of them could be neutral (more generic and task-based).} We then determine which one to retrieve as a response for each input context, based again on the TF-IDF vector similarity method described above.
\section{Experimental Setup}
\label{sect:Experimental Setup}
\subsection{Datasets}
\label{subsect:datasets}
As discussed above, we propose models that can deal with style data coming from an unpaired, non-parallel domain, different from the domain of the dialogue dataset.
For our style (politeness) domain, we use the \textit{Stanford Politeness Corpus}~\cite{Danescu-niculescu-mizil2013}, which contains a collection of requests from \textit{Wikipedia} (WIKI) editor's talk pages
and the \textit{Stack Exchange} (SE) question-answering communities.
Based on scores from human annotators, 
these requests are labeled with either polite or rude, 
with each class equally consisting of 1,089 requests for the 
Wikipedia domain and 1,651 requests for the 
Stack Exchange domain.
For the content (dialogue) domain, we use the popular \textit{MovieTriples} dialogue corpus~\cite{serban2016building}, which contains 245K conversations extracted from IMSDB movie scripts in \textit{X-Y-X} triplet-utterance format, where X and Y correspond to two movie characters (and the model's task is to generate the last response).

\subsection{Evaluation Methods}
\label{subsect:Evaluation Methods}
\paragraph{Human} To evaluate our models' ability to generate
polite responses without sacrificing dialogue quality,
we conducted several comprehensive human evaluation studies
on Amazon Mechanical Turk (MTurk).
Specifically, we compare the three stylistic models 
w.r.t. the base model on both dialogue quality 
(i.e., context relevance and coherence) and politeness level.\footnote{We opted for dialogue quality rather than several separated, fine-grained metrics such as relevance, specificity, informativeness because~\newcite{lowe2017towards} found that little additional information was provided by adding in more metrics on top of overall dialogue quality, and it also confused MTurkers in many scenarios. We had similar observations in our initial human study on MTurk.}
For this, we randomly sampled $300$ contexts covering all types of conversations and their generated responses from the Seq2seq base model, 
the three stylistic models, 
and the retrieval-based models.
For each source input, the six responses are randomly shuffled to anonymize model identities.
Each response was then annotated by two human evaluators that were located in the US, had an approval rate greater than $98\%$, and had at least $10,000$ approved HITs on record (to prevent those who had just started using MTurk and hence unconditionally enjoyed a high acceptance rate.).
All our human evaluations are performed by two annotators (for both dialogue quality and politeness level) in order to calculate inter-rater agreement,
for which we employ Cohen’s Kappa $\kappa$~\cite{c68}, a score that measures the level of inter-rater agreement between two annotators on a classification problem~\cite{artstein2008inter}.
For both dialogue quality and politeness evaluations, 
the human raters were shown the conversation context (input) 
and the six shuffled responses (from the six models).
Clear instructions (closely following those from~\newcite{wang2017steering}) corresponding to each score were shown in the interface.
More specifically, we asked the annotators to first read the context and each of the generated/retrieved responses, and assign to each response a score.
They then scored each response on a five-point Likert
scale~\cite{likert1932technique} (for both politeness and dialogue quality), 
hence providing absolute measurements but in an overall comparative (relative)
setting.\footnote{The Likert
scale is a bipolar scaling method that maps each score to a text item
that describes the score, e.g., our politeness level interface uses `Polite', `Slightly Polite', `Neutral', `Slightly Rude', `Rude'; and our dialogue quality study uses `Very good',
`Good', `Acceptable', `Poor', and `Very poor', instead of the abstract scores $1$-$5$. 
Note that we did not adopt pairwise comparisons because first, 
it will create several independent sets of pairwise results ($15$ sets in our case), which also raises the cost substantially, and secondly, pairwise comparison does not tell us "by how much" a response is better/equal/worse than the other. In contrast, our absolute scores can help future research compare more directly to our results.
We will release our detailed instructions and MTurk interfaces, plus our annotation scores on a public webpage.}
We explicitly stated that it is possible for them to find some conversation disconnected or lacking context, and encouraged them to make the best guess when in doubt.
Using similar instructions (and a $300$-sized sample), we also performed a separate 3-way LFT model comparison by setting its target politeness scores to $1.0$, $0.5$ and $0.0$, respectively.

\paragraph{Automatic} Since there do not exist ground-truth stylized versions of the response to the MovieTriples conversations, we only use automatic evaluation metrics as complementary and trend-verification information to the primary human perception studies in this work: we compute BLEU (a phrase-matching based metric;~\cite{Papineni:2002:BMA:1073083.1073135}) as an approximation of dialogue quality as used by some previous work~\cite{Ritter:2011:DRG:2145432.2145500,galley2015deltableu,DBLP:journals/corr/LiGBGD16}.
Note that we choose to report BLEU scores not in order to draw any immediate conclusion (\newcite{liu2016not} found that BLEU does not correlate well with human studies on dialogue quality), but rather to check for match with the trends from human evaluation.
We also compute the politeness classifier's scores as an approximation of politeness level.
Sec.~\ref{sec:style-eval} discusses these results.

\subsection{Training Details}
\label{sec:trainingdetails}
We now present some important training details.\footnote{We will add all reproducibility details and more analysis examples in a post-publication supplement on our webpage.}

\paragraph{Embedding Initialization}
For all our models, we initialized the embedding matrix 
with word2vec trained on Google News dataset (about 100 billion words)\footnote{\url{https://code.google.com/archive/p/word2vec/}}~\cite{DBLP:journals/corr/abs-1301-3781}; we use \textit{Xavier} initializer~\cite{Glorot10understandingthe} for out-of-vocabulary words.

\paragraph{Pretraining}
Following~\newcite{serban2016building},
we pretrained the Seq2seq base model for $4$ epochs 
with Q-A SubTle corpus~\cite{ameixa2014luke}, 
which contains around $5.5$M movie subtitle Q\&A pairs. 

\paragraph{Implementation Details}
We used 300-dim embeddings, the \textit{AdamOptimizer}~\cite{kingma2014adam} 
with a learning rate of $0.001$, and a dropout rate of $0.2$.
All models were trained with a mini-batch of size $96$.
The classifier was trained for 3 epochs,
and the three proposed stylistic models were each
trained for 35 epochs. The polite language model 
used in the Fusion model was trained until there was no improvement 
for perplexity on a held-out dev-set (all tuning decisions were made on the respective dev-sets).
We use a balanced value of $0.5$ for the fusion ratio ($\alpha$ in Eq.~\ref{eq:fusion}), and $2.0$ for the RL weight ($\beta$ in Eq.~\ref{eq:reinforce}) after some light empirical tuning.
Also due to the nearly perfect balance
between the number of polite and rude examples
in the Stanford Politeness Corpus,
we set the baseline reward of Polite-RL ($R_b$ in Eq.~\ref{eq:reinforce}) to a constant $0.5$ at all times.\footnote{We also tried using a self-critical baseline as in~\newcite{rennie2016self}, but found that our way of setting the constant-based baseline led to better responses.
We speculate that this is because a self-critical approach tries to make an utterance as polite as possible, which usually leads to a few very generic and very polite responses at convergence (because the model gets a positive reward only when the sampled utterance is more polite than the greedy-decoded one).}
Note that for effective and non-confusing MTurk studies, 
for all our models (the base model and the three stylistic models), 
we avoid UNK tokens to appear in the generated response, 
by not back-propagating the MLE loss for these tokens. 
We also do the same for a short list (around 10) of very offensive swear words (from Wiktionary).

\section{Results}
\label{sect:Results}
In this results section, we first briefly present our politeness classifier (Sec.~\ref{sect:Politeness Classification Model}) and base dialogue model (Sec.~\ref{subsect:Base Seq2seq Dialogue Model}) results, and then focus on the stylistic-dialogue results (retrieval and generative).

\begin{table}[t]
\begin{small}
  \centering
    \begin{tabular}{lcc}
    	\toprule
          & WIKI  & SE \\
    SVM   & 82.6\% & 65.2\% \\
    CNN   & \textbf{85.8\%} & 66.4\% \\
    LSTM-CNN \ \ \ \ \  \  & 85.0\% & \textbf{70.2\%} \\
    	\bottomrule
    \end{tabular}%
    \vspace{-4pt}
  \caption{Politeness classification accuracies. Top results are boldfaced.}
  \vspace{-6pt}
  \label{tab:classification-accuracy}%
\end{small}
\end{table}%

\subsection{Politeness Classification Results}
\label{subsect:Politeness Classification Results}
Following~\newcite{Danescu-niculescu-mizil2013},
we use accuracy (i.e., percentage of correctly labeled messages for binary polite/rude labels) to evaluate our politeness classifier's generalization ability.
Specifically, we used data from the training set of WIKI,
and test on both the test set of WIKI and the 
entire SE (StackExchange) corpus. We used the same train-validation-test split setup 
($7$:$1$:$2$) as in~\newcite{Aubakirova2016}.\footnote{Note that this train/dev/test split is only 
for verifying the strength of the classification model. 
The classifier used for the three proposed polite-dialogue models was trained on the entire Stanford Politeness Corpus (due to the small amount of politeness-labeled data available).}
As shown in Table~\ref{tab:classification-accuracy}, our LSTM-CNN model improved cross-domain accuracy (while maintaining comparable in-domain accuracy) compared to that of the SVM and CNN models reported in~\newcite{Aubakirova2016}. This is similar to how~\newcite{zhou2015c} also found that a combination of LSTM-RNNs and CNNs is superior to an LSTM-RNN or CNN alone for sentiment classification, likely because the joint model captures both long-distance relationships as well as local windowed filter-based features, and this could make it easier to separate in-domain and out-of-domain properties. Also, we observe more improvement on cross-domain accuracy because it has much more space for improvement, as opposed to in-domain accuracy which is already very close to human performance.
The higher accuracy is also important because we need a cross-domain-accurate style classifier so that it can effectively stylize responses in diverse dialogue corpora domains such as MovieTriples.

\begin{table}[t]
\begin{small}
  \centering
    \begin{tabular}{lcccc}
    \toprule
    Model & \multicolumn{1}{l}{PPL} & \multicolumn{1}{l}{PPL@L} & \multicolumn{1}{l}{WER} & \multicolumn{1}{l}{WER@L} \\
    RNN   & 27.09 & 26.67 & 64.10 & 64.07 \\
    HRED  & 27.14 & 26.60  & 64.10 & 64.03 \\
    HRED-Bidir. & 26.81 & 26.31 & \textbf{63.93} & \textbf{63.91} \\
    Seq2seq & \textbf{25.96} & \textbf{25.85} & 64.27 & 64.25 \\
    \bottomrule
    \end{tabular}%
    \vspace{-5pt}
  \caption{PPL, WER results computed on $\{U_1, U_2, U_3\}$ and PPL@L, WER@L computed on $\{U_3\}$ conditioned on $\{U_1, U_2\}$. Lower is better for all metrics. Top results are boldfaced.}
  \vspace{-6pt}
  \label{tab:ppl+wer}%
  \end{small}
\end{table}%

\subsection{Base Dialogue Model Results}
\label{subsect:Base Dialogue Model Results}
Next, in Table~\ref{tab:ppl+wer}, we show that our starting point, base dialogue model is comparable in quality to a popular, representative previous model of~\newcite{serban2016building}, trained on the same corpora with similar model architectures.
We use their \textit{Perplexity} (\textit{PPL}) and \textit{Word Error Rate} (\textit{WER}) metrics.
In order to have a meaningful perplexity (i.e., the probability of regenerating a reference response) comparison between two language generation models, 
they should have the same vocabulary set.
Since the vocabulary of our politeness dialogue models is a combination of vocabulary sets drawn from the MovieTriples and Stanford Politeness corpora, for fair comparison in this section, we separately train a base Seq2seq model following exactly the vocabulary ($10,000$ most frequent tokens, plus an UNK for the rest) and preprocessing protocols from~\newcite{serban2016building}.
We bootstrapped the model with $4$ epochs
on the SubTle corpus (see Sec.~\ref{sec:trainingdetails}), and then trained on MovieTriples till there was no improvement on perplexity for the validation set.
The comparison for this base model with their hierarchical-encoder HRED models is presented
in Table~\ref{tab:ppl+wer}. As shown, we get comparable results overall on all metrics, and hence we have a good starting-point dialogue model to next add politeness to via three approaches.

\subsection{Stylistic Dialogue Model Results}
\label{sec:style-eval}
\paragraph{Primary Human Evaluation Results}
In this section, we present our primary human evaluation (MTurk) results
on both politeness level and dialogue quality (context-relevance) of the generated response, based on two annotators and a 300-sized test sample.
Table~\ref{tab:polite-result} shows the annotator-average scores for each of these two metrics and their absolute difference, based on our Likert scales of $1$ to $5$ (see Sec.~\ref{subsect:Evaluation Methods}).
We can first see that all three of our stylistic generative models improve on politeness compared to the Seq2seq base model. 
However, the Fusion model's politeness gain is not statistically significant,\footnote{We test stat. significance via the bootstrap test~\cite{noreen1989computer,efron1994introduction} with 100K samples.} and moreover it achieves this minor politeness level improvement at the cost of significantly compromising dialogue quality (because
its output is half-determined by a standalone politeness-trained LM that ignores context).

Next, we see that the LFT model is the most polite
(stat. significance of $p<0.01$ over the Seq2seq model), 
and also has dialogue quality close (statistically equal) to that of Seq2seq.
Our final Polite-RL model wins over Seq2seq on politeness (stat. significance of $p<0.01$) as well as achieves a small improvement in dialogue quality (though not at stat. significance level; but it is stat. significantly better in quality than Retrieval, Generic-10 and Fusion.). Moreover, the politeness levels of the LFT and Polite-RL models are statistically equal. Therefore, both models, with their training depth and multitasking trade-offs (see Sec.~\ref{sect:polite-dialogue-models}), can produce strong levels of stylistic content, without harming context-relevance.

\begin{table}[t]
  \centering
  \scalebox{0.85}{
    \begin{tabular}{l | cc | c}
    \toprule
          & Politeness & Quality & Difference \\ 
    Retrieval & 3.57 & 3.15 & 0.42\\ 
    Generic-10 & \textbf{3.66} & 2.99 & 0.67\\
    Seq2seq & 3.11  & 3.42 & 0.31\\
    Fusion & 3.23  & 3.05 & 0.18\\
    LFT   & 3.63  & 3.39 & 0.24\\
    Polite-RL & 3.50 & \textbf{3.54} & \textbf{0.04}\\
    \bottomrule
    \end{tabular}%
  }
  \vspace{-5pt}
\caption{MTurk human evaluation results on politeness level and dialogue quality (as well as the absolute value difference between the two, to show balance) of the Retrieval Models, Seq2seq and the three proposed generative models (avg. of two annotators is shown here). Top results are boldfaced.}
\label{tab:polite-result}%
\end{table}%

Lastly, we can also see that our two retrieval-based models are both very polite (but not stat. significantly better over LFT); and as expected, they both have dialogue quality lower than Seq2seq, Polite-RL and LFT (stat. significance of $p<0.01$). 
They also feature two of the worst balances between average politeness and dialogue quality score. 
This is the type of sacrifice we want to avoid from imposing on dialogue quality when building a stylistic dialogue model.

For inter-annotator agreement, the Kappa score was $0.35$ (fair\footnote{These levels were defined by~\newcite{landis1977measurement}; also see {\scriptsize\url{https://en.wikipedia.org/wiki/Cohens_kappa}}}) on Dialogue Quality and $0.46$ (moderate) on Politeness.
If we employ a collapsed-Likert version, where the more ambiguous and extreme scores of $\{1, 2\}$ and  $\{4, 5\}$ are bucketed together,\footnote{As discussed in~\newcite{weijters2010effect},~\newcite{james1984estimating}, and {\scriptsize\url{https://en.wikipedia.org/wiki/Likert_scale}}, the `central tendency bias' makes raters avoid using the two extreme response categories.} we obtained a Kappa score of $0.42$ (moderate) on Dialogue Quality and $0.55$ (moderate) on Politeness.

\paragraph{Human Evaluation Results on 3-way LFT Models}
We also present results on a 3-way politeness level comparison MTurk study among the Polite-LFT, Neutral-LFT,
and Rude-LFT models, i.e., the LFT model with three levels (scores) of scaling the prepended style label, corresponding to politeness scores $1.0$, $0.5$ and $0.0$, respectively (Table.~\ref{tab:LFT-3-way}, \textit{Continuous-LFT} column). The table shows that Polite-LFT is significantly more polite than Neutral-LFT 
(stat. significance of $p<0.01$), 
and Neutral-LFT is in turn more polite than Rude-LFT
(stat. significance of $p<0.01$).
For inter-annotator agreement on this 3-way LFT study, we get a Kappa of $0.51$ (moderate), and $0.61$ (substantial) for the collapsed-Likert case.

We also experimented earlier with a discrete version of LFT, where we treated responses in the $[0.8, 1.0]$ range as \textit{polite}, $[0.2, 0.8]$ as \textit{neutral}, and $[0.0, 0.2]$ as \textit{rude}. Instead of scaling a single label embedding with continuous politeness scores (as described in Section~\ref{subsect:Label-Fine-Tuning (LFT)}), we assigned to each response one of these three
labels with no scaling, according to its corresponding politeness bin.
The human evaluation scores for that model were $3.52$, $3.09$ and $2.93$, respectively, 
which features less score difference between \textit{neutral} and \textit{rude} (Table.~\ref{tab:LFT-3-way} \textit{Discrete-LFT} column).

\begin{table}[t]
\begin{small}
  \centering
    \begin{tabular}{lcc}
    \toprule
          & Continuous-LFT & Discrete-LFT\\
    Polite & 3.70 & 3.52\\
    Neutral & 3.15 & 3.09\\
    Rude & 1.19 & 2.93\\
    \bottomrule
    \end{tabular}
    \vspace{-2pt}
  \caption{MTurk human evaluation results on politeness level of 3 LFT models, for both the continuous and the discrete versions.}
  \label{tab:LFT-3-way}%
  \end{small}
\end{table}%

\paragraph{Automatic Metric Evaluation Results}
As discussed in Sec.~\ref{subsect:Evaluation Methods}, we also use some automatic evaluation metrics to complement and verify the MTurk human study results. 
In Table~\ref{tab:auto-metrics}, we present the average politeness classifier and BLEU-4 scores of responses from each model.
First, we can see that our politeness classifier agrees reasonably well with the human politeness judgments in Table~\ref{tab:polite-result}, since both identify the Retrieval-based models and LFT as the most polite, followed by Polite-RL and Fusion in descending order. 
We quantified this `agreement' concretely, and found high correlation between the six human Politeness scores (Table~\ref{tab:polite-result} \textit{Politeness} column) and the six automatic classifier scores (Table~\ref{tab:auto-metrics} \textit{Politeness Score} column): Pearson correlation is $0.827$ (stat. significance $p=0.0422$), and Spearman's rank-order correlation is $0.9276$ ($p=0.0077$).
Next, for BLEU scores, although these scores (as percentages) are very low (consistent with the observation in~\newcite{Ritter:2011:DRG:2145432.2145500} and~\newcite{DBLP:journals/corr/LiGBGD15}), 
their relative system-ranking still roughly agrees with that of human judgments --- we found reasonably high correlation between human Dialogue Quality and BLEU (based on the six scores in Table~\ref{tab:polite-result} \textit{Quality} column and Table~\ref{tab:auto-metrics} \textit{BLEU-4} column): Pearson correlation is $0.793$ (stat. significance $p=0.0597$), and Spearman's rank-order correlation is $0.771$ ($p=0.0724$).

Hence, overall, the automatic metric evaluation again shows that without
politeness training, the base dialogue model produces neutral responses on average (0.49 score), while the retrieval-based models and all three proposed generative models improve on politeness score.
Also, the BLEU scores show, similar to the human study results in Table~\ref{tab:polite-result}, 
that among the three proposed models, the Fusion model sacrifices the most dialog quality to become more polite, 
whereas the LFT and RL models maintain comparable quality with improved politeness over the base model (Seq2seq). 
For the retrieval models, we again see that their politeness levels are better than LFT and RL models, 
but with a corresponding loss in dialogue quality.

\begin{table}[t]
\begin{small}
  \centering
    \begin{tabular}{lcc}
    \toprule
          & Politeness Score & BLEU-4 \\
    Retrieval & 0.88 & 0.59 \\
    Generic-10 & \textbf{0.93} & 0.03 \\
    Seq2seq & 0.49 & \textbf{1.05} \\
    Fusion & 0.61 & 0.78 \\
    LFT   & 0.72 & 1.02 \\
    Polite-RL & 0.61 & 0.94 \\
    \bottomrule
    \end{tabular}
    \vspace{-5pt}
  \caption{Automatic metrics: avg. politeness and BLEU-4 scores for the two Retrieval models, Seq2seq and three proposed models. Also, the politeness score of Neutral-LFT and Rude-LFT are 0.48, 0.25, resp. Top results are boldfaced.}
  \vspace{-6pt}
  \label{tab:auto-metrics}%
  \end{small}
\end{table}%

\begin{table}[t]
\begin{footnotesize}
  \centering
    \begin{tabular}{|p{6.3cm}|r|}
    \hline
    \textbf{Target Sequence} & \textbf{Score} \\
    \hline
    \hline
    \multicolumn{2}{|c|}{Polite Examples} \\
    \hline
    well , thanks . that 's . i appreciate that . & 0.99 \\
    $\langle$num$\rangle$ , $\langle$num$\rangle$ of them in los angeles . i checked . nice work , though . & 0.98 \\
    nah . i have curfew . he starts to walk away , then stops . quincy oh , by the way . congratulations . & 0.97 \\
    thank you , ma'am . um , may i ask what this is regarding ? & 0.91 \\
    hi , $\langle$person$\rangle$ . how are you ?  & 0.89 \\
    i know . amazing . & 0.89 \\
    \hline
    \hline
        \multicolumn{2}{|c|}{Rude Examples} \\
    \hline
    you really should pay more attention to what you read , $\langle$person$\rangle$ . & 0.11 \\
    they were in a car accident . & 0.10 \\
    you calling more of your stupid friends again ? ya prick . & 0.05 \\
    are you a bad boy ? & 0.05 \\
    then she 's a liar . & 0.04 \\
    excuse me , does that flask belong to this man ? & 0.03 \\
    oh , well , excuse me all to hell . & 0.02 \\
    \bottomrule
    \end{tabular}
    \vspace{-5pt}
  \caption{Analysis of politeness classifier: sampled responses from MovieTriples dialogue corpus (without cherry-picking) and their politeness score. The double line separates polite and rude classified responses.\vspace{-10pt}}
  \label{tab:label-error-analysis}%
  \end{footnotesize}
\end{table}%

\begin{table}[ht!]
\begin{footnotesize}
  \centering
    \begin{tabular}{lp{5.4cm}}
    \toprule
    \textbf{Input} & X: for what ? \\
          & Y: for forcing me to face my life honestly for the first time . \\
    \textbf{  Retrieval} & X: what 's the c . w . for ? \\
    \textbf{  Generic-10} & X: thanks for the answer . \\
    \textbf{  S2S} & X: that 's not true . \\
    \textbf{  Fusion} & X: i do n't know . I mean , it 's just a feeling , but that 's not what i 'm saying . \\
    \textbf{  LFT} & X: you 'll be fine . \\
    \textbf{  RL} & X: i wish i could . \\
\midrule
    \textbf{Input} & X: no, but i wanted to. \\
          & Y: are you a total lying , scumsucking pig , $\langle$person$\rangle$ ? \\
    \textbf{  Retrieval} & X: so $\langle$person$\rangle$ , what did it total out at ? \\
    \textbf{  Generic-10} & X: no problem . \\
    \textbf{  S2S} & X: oh , that 's ridiculous . \\
    \textbf{  Fusion} & X: well , you 're not a little funny . \\
    \textbf{  LFT} & X: i do n't think so , sir . \\
    \textbf{  RL} & X: you 're hurting me ! \\
\midrule
    \textbf{Input} & X: you 're sweet to say so . \\
          & Y: pretty song . \\
    \textbf{  Retrieval} & X: is there a song you do n't know ? \\
	\textbf{  Generic-10} & X: can you clarify ? \\
    \textbf{  S2S} & X: $\langle$person$\rangle$ , i know . \\
    \textbf{  Fusion} & X: you 're a good man . \\
    \textbf{  LFT} & X: i love this song . \\
    \textbf{  RL} & X: you sound like a goddess . \\
\midrule
    \textbf{Input} & X: i find it hard to believe myself . \\
          & Y: his lips , unused as they are to such a contortion , twisted and stretched . into a smile . \\
    \textbf{  Retrieval} & X: thank you . and always smile . \\
	\textbf{  Generic-10} & X: nice . \\
    \textbf{  S2S} & X: how can you be so sure ?
 \\
    \textbf{  Fusion} & X: you 're a good man , mr . \\
    \textbf{  LFT} & X: your lips are well . \\
    \textbf{  RL} & X: your lips are so beautiful . \\
    \bottomrule
    \end{tabular}%
    \vspace{-5pt}
    \caption{Output dialogue response examples by Retrieval, Generic-10, Seq2seq (denoted as S2S) and the 3 generative polite models Fusion, LFT, and RL (shows conversation history turns of X and Y, and then the generated 3rd turn response by X).}
    \vspace{-10pt}
  \label{tab:outputexamples}%
  \end{footnotesize}
\end{table}%

\section{Analysis}
\label{sect:Analysis}
\subsection{Analysis of Politeness Classifier}
\label{subsect:Qualitative Analysis of the Classifier on MovieTriples}
We start our analysis by providing qualitative examples of 
how well our politeness classifier performs on the target sequences from MovieTriples train dataset. This is important to check because the classifier is trained on Wikipedia (Wiki) admin request messages, and while our LSTM-CNN performs better on cross-domain StackExchange (SE) data, the MovieTriples dialogue corpus is still quite different and diverse in domain from both Wiki and SE. Hence, it is important to have a reasonably accurate politeness classifier such that it can provide useful labels and rewards for our polite-dialogue models.
Table~\ref{tab:label-error-analysis} presents 
some randomly-selected (i.e., non-cherry-picked) responses from MovieTriples and their politeness classifier scores.
We can see that the classifier provides a reasonably correct score a majority of the time, capturing several psycholinguistic politeness strategies mentioned in~\newcite{Danescu-niculescu-mizil2013}, e.g., positive ones such as gratitude, deference, greeting, positive lexicon, indirection, indicative modal, and negative ones such as negative lexicon, direct question, direct start, 2nd person start.
However, it does occasionally give strongly polite or rude scores to some mild or neutral responses, e.g., "they were in a car accident", showing scope for classifier improvements.

\begin{figure}[t]
\centering
\includegraphics[width=0.48\textwidth]{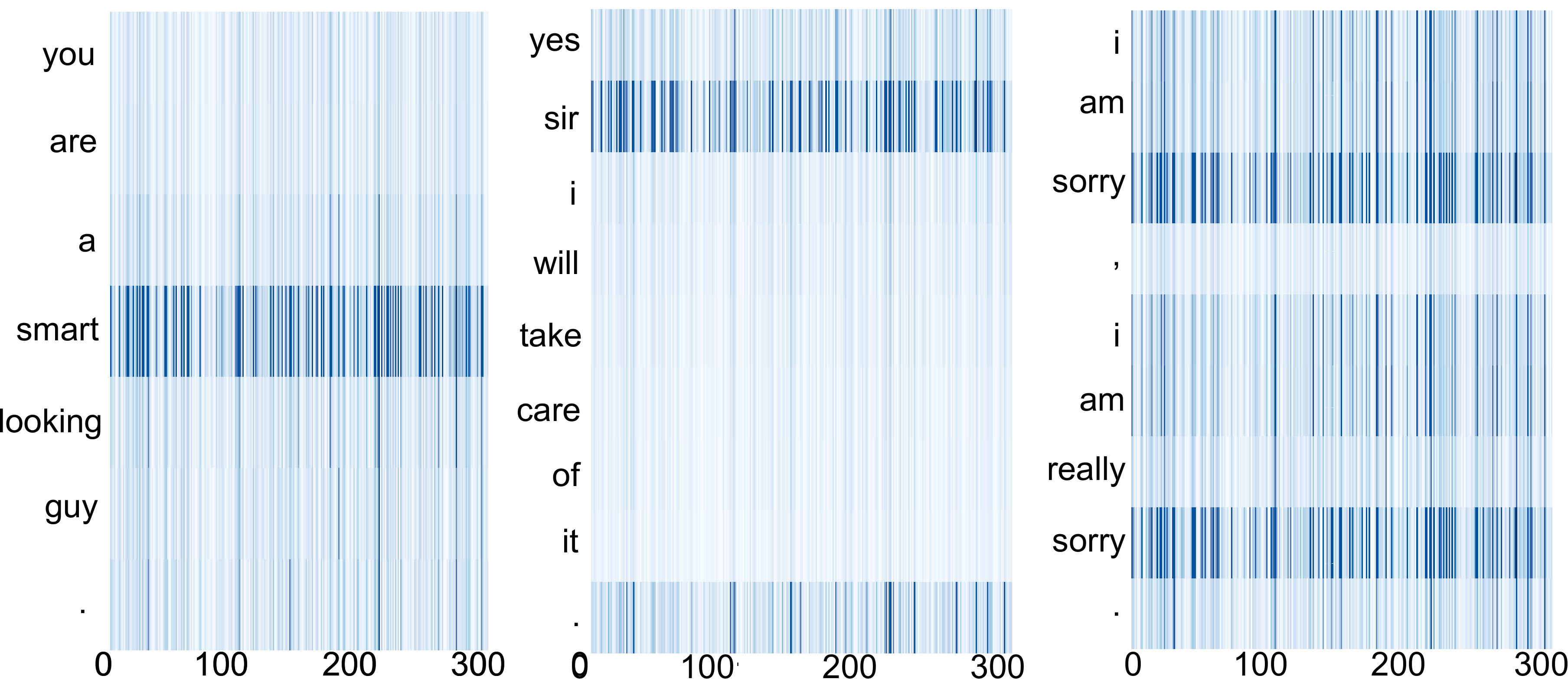}
\vspace{-20pt}
\caption{Saliency heatmaps of the classifier's attention (reward for sampled responses in Polite-RL model).}
\vspace{-10pt}
\label{fig:vis}
\end{figure}

\vspace{-5pt}
\subsection{Output Examples of Stylistic Dialogue}
\vspace{-2pt}
\label{subsect:Output Examples}
Next, we show some output examples of our polite dialogue models w.r.t. the base Seq2seq model as well as the retrieval-based models. We use these examples to demonstrate the politeness strategies our proposed generative models have learned (in Table~\ref{tab:outputexamples}).
In the first example, our stylistic models
use politeness strategies such as 
indirection, positive lexicon
and counterfactual modal~\cite{Danescu-niculescu-mizil2013}.
This example also illustrates the behavior of the Retrieval model, i.e., most of the time it just outputs an utterance that has word overlap with but totally irrelevant to the context.
Thus although all its retrieved responses have oracle-level fluency and grammaticality, its average dialogue quality score in the human evaluation is still not as good as that of Seq2seq.

In the second example, Fusion uses indirection,
while LFT is being polite even when disagreeing with the abusive language from $Y$. This example also shows that Generic-10, due to its limited space for retrieval, oftentimes fails to provide a relevant answer, although it is the most polite one since its candidate responses are manually picked.
In the third example, Fusion and LFT both use positive lexicon, and RL makes a compliment.
In the fourth example, each of the three proposed models uses positive lexicon.
It is worth noting that in the last example, 
while LFT and Polite-RL seem to provide a relevant 
compliment, they are actually complimenting
the wrong person. This kind of issue
motivates us toward creating persona-based~\cite{DBLP:journals/corr/LiGBGD16}
politeness models for future work.

\subsection{Visualization of Polite-RL Reward}
Using derivative saliency~\cite{DBLP:journals/corr/SimonyanVZ13,li2015visualizing,Aubakirova2016}, we also visualize how much each token in the sampled response contributes to the classifier's reward during Polite-RL model's training.
Fig.~\ref{fig:vis} shows three such heatmaps 
that correspond to the magnitudes of the derivative 
in absolute value with respect to each dimension.
The figures clearly show that the classifier has learned 
to identify multiple politeness strategies, e.g., "smart" (deference), "sir" (polite address), and the two "sorry"s (apologizing).

\section{Conclusion and Future Work}
\vspace{-5pt}
\label{sect:Conclusion and Future Work}
We first presented three diverse generative models that can generate rich polite-to-rude spectrum dialogue responses (based on the politeness theories by~\newcite{brown1987politeness}), 
without using any parallel data (which is usually assumed for tasks such as machine translation) and only relying on a style classifier. 
Via multiple human evaluation studies and automatic metrics, we demonstrated that all three models generate more polite responses (displaying several politeness strategies discussed in previous psycholinguistic works), while LFT and Polite-RL are able to do so without losing dialogue quality, as opposed to the Fusion model as well as the two retrieval-based models.

In future work, there is still much room for improvement on the politeness as well as dialogue quality side, and one could employ more recent, advanced models such as variational, adversarial, and decoder-regulation techniques.

Though we focused on politeness for the scope of this paper, our models can be easily generalized to other emotion and personality styles (only relying on a style classifier), hopefully contributing towards the valuable paradigm of human-like and engaging intelligent tutors and personal assistants. 
In future work, our polite-RL model could also be extended to stylistic task-based dialogue generation, 
where both content preservation and style transfer are needed, 
potentially by disentangling politeness and content of the
generated response and then only feeding the politeness portion to the classifier for RL training.
\section*{Acknowledgments}
We thank the action editor and the anonymous reviewers for their helpful comments and discussions. This work was supported by DARPA
(YFA17-D17AP00022), Facebook ParlAI Research Award, Google Faculty Research Award, Bloomberg Data Science Research Grant, and Nvidia GPU awards. The views, opinions, and/or findings contained in this article are those of the authors and should not be interpreted as representing the official views or policies, either expressed or implied, of the funding agency.

\bibliography{references}
\bibliographystyle{acl2012}

\end{document}